\documentclass[letterpaper]{article} 
\usepackage{aaai24}  
\usepackage{times}  
\usepackage{helvet}  
\usepackage{courier}  
\usepackage[hyphens]{url}  
\usepackage{graphicx} 
\usepackage{natbib}  
\usepackage{caption} 
\usepackage{multicol}
\usepackage{multirow}
\usepackage{booktabs} 
\usepackage{xcolor}
\usepackage{colortbl}
\usepackage{enumitem}
\usepackage{amsmath}
\usepackage{amsthm}
\usepackage{bm}
\usepackage{pifont}
\usepackage{amsfonts}
\usepackage[ruled]{algorithm2e}

\definecolor{aliceblue}{rgb}{0.94, 0.97, 1.0}

\newcommand{\ie}{\textit{i}.\textit{e}.}
\newcommand{\eg}{\textit{e}.\textit{g}.}
\usepackage{newfloat}
\usepackage{listings}
\DeclareCaptionStyle{ruled}{labelfont=normalfont,labelsep=colon,strut=off} 
\title{Exploiting Auxiliary Caption for Video Grounding}
\author {
    Hongxiang Li\textsuperscript{\rm 1},
    Meng Cao\textsuperscript{\rm 2},
    Xuxin Cheng\textsuperscript{\rm 1},
    Yaowei Li\textsuperscript{\rm 1},
    Zhihong Zhu\textsuperscript{\rm 1},
    Yuexian Zou\textsuperscript{\rm 1}\footnotemark[2]
}
\affiliations {
    \textsuperscript{\rm 1}School of Electronic and Computer Engineering, Peking University\\
    \textsuperscript{\rm 2}International Digital Economy Academy (IDEA)\\
    \{lihongxiang, chengxx, ywl, zhihongzhu\}@stu.pku.edu.cn; \{mengcao, zouyx\}@pku.edu.cn
}

\usepackage{bibentry}

\begin{document}

\maketitle

\renewcommand{\thefootnote}{\fnsymbol{footnote}}

\footnotetext[2]{~Corresponding author.}
\begin{abstract}
Video grounding aims to locate a moment of interest matching the given query sentence from an untrimmed video. Previous works ignore the {sparsity dilemma} in video annotations, which fails to provide the context information between potential events and query sentences in the dataset. In this paper, we contend that exploiting easily available captions which describe general actions, i.e., auxiliary captions defined in our paper, will significantly boost the performance. To this end, we propose an Auxiliary Caption Network (ACNet) for video grounding. Specifically, we first introduce dense video captioning to generate dense captions and then obtain auxiliary captions by Non-Auxiliary Caption Suppression (NACS). To capture the potential information in auxiliary captions, we propose Caption Guided Attention (CGA) project the semantic relations between auxiliary captions and query sentences into temporal space and fuse them into visual representations. Considering the gap between auxiliary captions and ground truth, we propose Asymmetric Cross-modal Contrastive Learning  (ACCL) for constructing more negative pairs to maximize cross-modal mutual information. Extensive experiments on three public datasets (i.e., ActivityNet Captions, TACoS and ActivityNet-CG) demonstrate that our method significantly outperforms state-of-the-art methods. 
\end{abstract}

\section{Introduction}
\label{sec:intro}

Video grounding~\cite{CTRLgao2017tall,2D-TANzhang2020learning,MMNwang2022negative,cao2022correspondence,LGImun2020local,anne2017localizing,cao2022locvtp,zhang2022unsupervised,cao2023iterative,li2023g2l} aims to identify the timestamps semantically corresponding to the given query within the untrimmed videos. It remains a challenging task since it needs to not only model complex cross-modal interactions, but also capture comprehensive contextual information for semantic alignment.

\begin{figure}[t]
  \centering
    \includegraphics[width=1.0\linewidth]{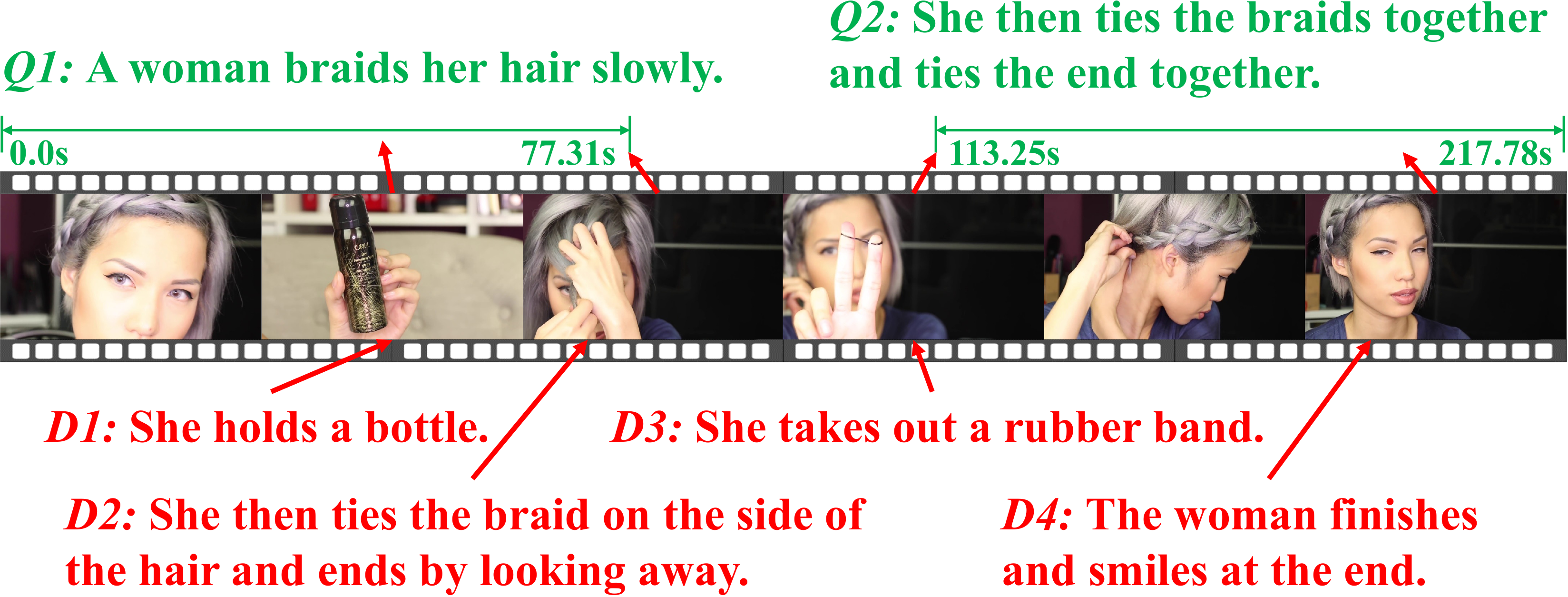}
   \caption{The sparse annotation dilemma in video grounding. The annotated captions (marked by {green}) in the dataset are sparse while there still exist many uncovered captions (marked by {red}). This 218-second video from ActivityNet Captions with 2 annotations.}
   \label{fig:intro}
\end{figure}

Currently, due to the costly labeling process, the annotations of existing video grounding datasets~\cite{krishna2017dense,regneri2013grounding} are \emph{sparse}, \ie, only a few actions are annotated despite the versatile actions within the video. For example in Figure~\ref{fig:intro}, the video from ActivityNet Captions~\cite{krishna2017dense} dataset lasts for 218 seconds and only 2 moment-sentence pairs (marked by green) are annotated. Previous methods ignore the presence of these unlabeled action instances (marked by red) associated with the query, which will facilitate the grounding. As shown in Figure~\ref{fig:intro}, the missing $D_3$ contains the process of ``\texttt{take out a rubber band}", which is preparatory for the action ``\texttt{tie the braids}" in the queried sentence $Q_2$.


\begin{figure}[t]
  \centering
   \includegraphics[width=0.9\linewidth]{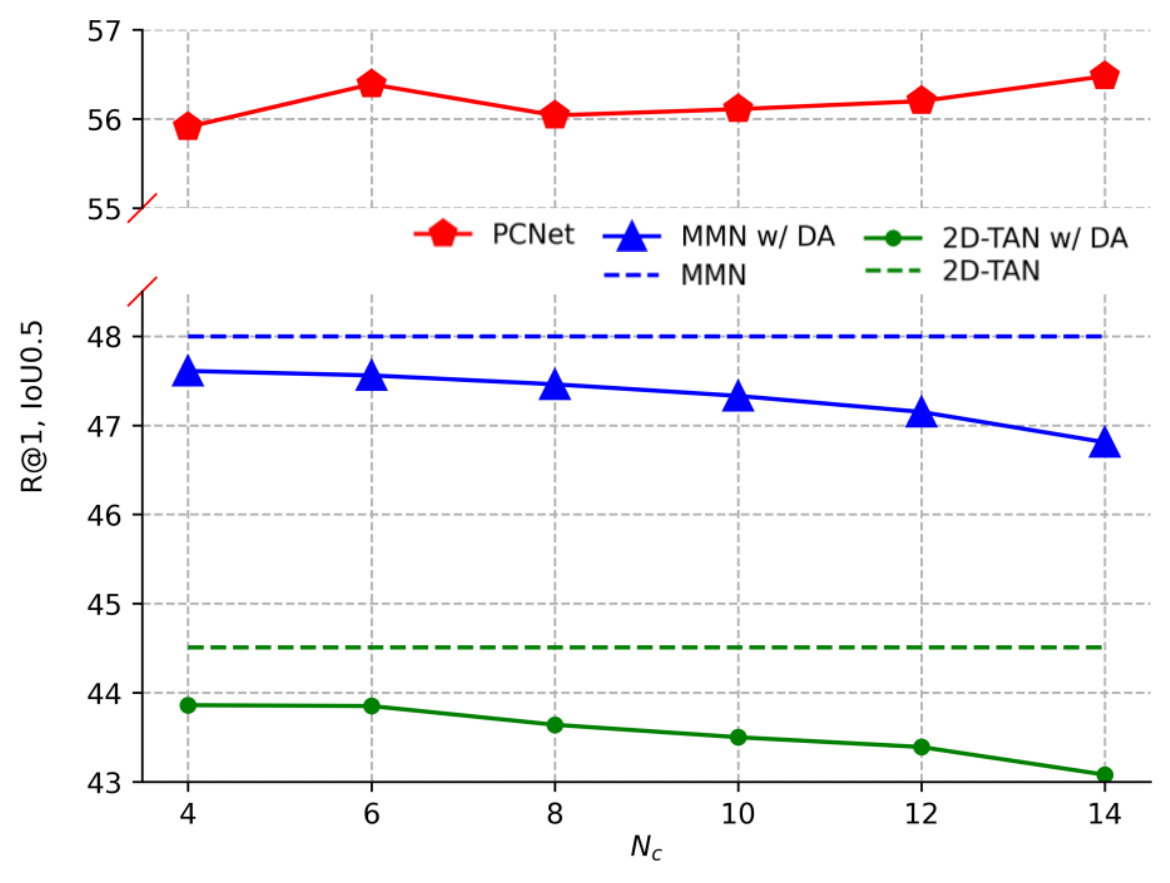}
   \caption{Performance comparison with ACNet and two representative models~\cite{MMNwang2022negative,2D-TANzhang2020learning} with dense caption data augmentation (w/ DA) on ActivityNet Captions. $l_c$ denotes the number of additional moment-sentence pairs per video.}
   \label{fig:com_da}
\end{figure}

However, it is labor-intensive to manually annotate all actions in the video. Recently end-to-end dense video captioning (DVC)~\cite{krishna2017dense,li2018jointly,suin2020efficient,PDVCwang2021end}, which combines event localization and video captioning together, has achieved satisfactory advances. A straightforward solution is to resort to dense video captioning for plausible caption generation. Intuitively, we can incorporate the DVC generated captions as a data augmentation (DA) strategy into the video grounding training. This simple solution, however, suffers from two inherent weaknesses: (1) The generated dense captions of timestamps and sentences may be rough and unreliable. (2) There may be overlaps between dense captions and ground truth. The incorrect caption of the ground truth moment will cause the model to learn incorrect information from training samples. Experimentally, we implement this data augmentation idea on two representative methods (\ie, MMN~\cite{MMNwang2022negative} and 2D-TAN~\cite{2D-TANzhang2020learning}). The experimental results on ActivityNet Captions dataset are shown in Figure~\ref{fig:com_da}. We have seen that directly using such data augmentation leads to performance degradation. For example, when using 10 additional dense captions, the performance drops by 1.01\% in 2D-TAN.

\begin{figure*}[t]
    \centering
    \includegraphics[width=1.\linewidth]{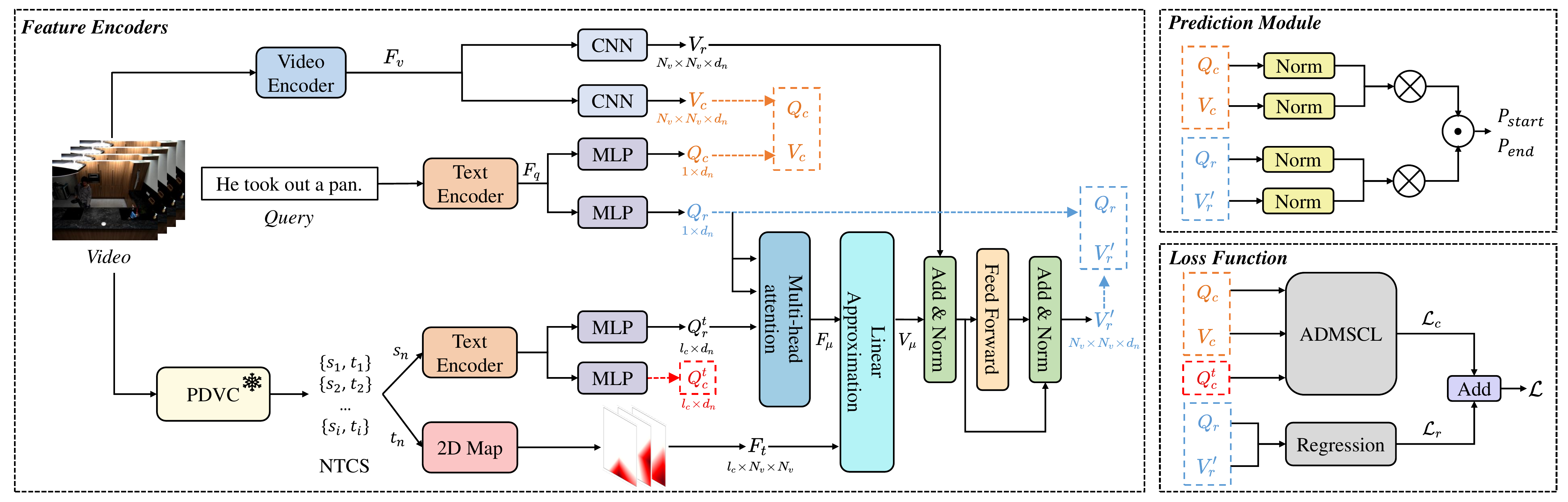}
    \caption{Overview of the proposed Auxiliary Caption Network (ACNet). Auxiliary Caption is filtered through our proposed Non-Auxiliary Caption Suppression algorithm (NACS) from PDVC~\cite{PDVCwang2021end} outputs. We convert the timestamp of the auxiliary caption to the 2D map form following~\cite{2D-TANzhang2020learning,MMNwang2022negative}. Then, video segments and query sentences are encoded by the respective feature encoders for regression learning and cross-modal contrastive learning. In the regression branch, Caption Guided Attention (CGA) calculates semantic relations between query features $Q_r$ and auxiliary caption features $Q_r^t$. Then we project them to visual space to obtain visual representations $V_r^\prime $ with prior knowledge. $V_r^\prime $ and query features $Q_r$ are used for prediction and loss computation. In the cross-modal learning branch, the encoded video features $V_c$ and query features $Q_c$ are directly fed into the prediction module and loss function. $\otimes$ and $\odot$ indicate matrix multiplication and Hadamard product, respectively.}
    \label{fig:pipeline}
\end{figure*}

Despite this intuitive data augmentation does not achieve improvements, we still argue that these dense descriptions contain beneficial information for video grounding. In this paper, we first generate several dense captions from the input video using the off-the-shelf dense video captioning model. To improve the reliability of the generated captions, we propose Non-Auxiliary Caption Suppression (NACS), which selects high-quality and general moment-sentence pairs from the dense captions, defined as auxiliary captions. Unlike the intuitive data augmentation strategy, we propose a novel Auxiliary Caption Network (ACNet) to maximize the utilization of the generated captions rather than simply as extra training data as shown in Figure~\ref{fig:pipeline}. Our ACNet exploits the potential information embedded in the auxiliary caption through the regression branch and contrastive learning branch.

In the regression branch, we propose Caption Guided Attention (CGA) to investigate the prior knowledge in the auxiliary caption. Our motivation lies in that the auxiliary caption is a well-established prior indication, \ie, it provides an approximate temporal range for the action needed to be grounded. Specifically, we obtain the correlation information between the sentence of the auxiliary caption and the input query through the cross-attention mechanism. Then, we encode the timestamp of the auxiliary caption into a two-dimensional temporal map and linearly project semantic relations into the temporal map to obtain visual features with prior knowledge. In this manner, video clips related to the query semantics are assigned higher weights and unrelated ones are assigned lower weights, providing prompt information for the subsequent localization module.

In the contrastive learning branch, we introduce Asymmetric Cross-modal Contrastive Learning (ACCL) to capture more negative samples in the auxiliary caption. Traditional cross-modal contrastive learning treats all classes equally~\cite{khosla2020supervised,wang2022partial}, which is exhibited in video grounding as matched fragment-sentence pairs are treated as positive pairs and mismatched fragment-sentence pairs are treated as negative pairs while pulling is applied within positive pairs and pushing among negative pairs. However, the generated auxiliary moment-sentence pairs are not as accurate as the manually annotated ones. Additionally, there exist conflicts between auxiliary caption and ground truth as they are independent of each other. Therefore, while pulling the ground truth pairs together, we push the auxiliary caption sentences away from the ground truth moments but do not push the auxiliary caption moments away from the ground truth sentences. Auxiliary captions provide more descriptions related to the video content, which are treated as hard negative pairs with the ground truth moments. Our ACCL mines more supervision signals from unannotated actions without compromising the original representation capability.

Our main contributions are summarized in three fields: 
\begin{itemize}[topsep=0pt, partopsep=0pt, leftmargin=13pt, parsep=0pt, itemsep=3pt]
	\item We present the sparse annotation dilemma in video grounding and propose to extract information about potential actions from unannotated moments to mitigate it.
	\item We propose Caption Guided Attention (CGA) to fuse auxiliary captions with visual features to obtain prior knowledge for video grounding. Moreover, we propose Asymmetric Cross-modal Contrastive Learning  (ACCL) to mine potential negative pairs.
	\item Extensive experiments on three public datasets demonstrate the effectiveness and generalizability of ACNet. 
\end{itemize}


\section{Related Work}

\noindent \textbf{Video Grounding.} Video grounding also known as natural language video localization and video moment retrieval, was first proposed by ~\cite{CTRLgao2017tall, anne2017localizing}. Existing methods are primarily categorized into proposal-based methods and proposal-free methods. Proposal-based methods focus on the representation, ranking, quality and quantity of proposals. They perform various proposal generation methods such as sliding windows ~\cite{CTRLgao2017tall, anne2017localizing,ning2021interaction}, proposal networks ~\cite{xiao2021boundary,chen2019semantic}, anchor-based methods ~\cite{chen2018temporally,CSMGANliu2020jointly,2D-TANzhang2020learning} to extract candidate moments and then semantically match a given query sentence with each candidate through multi-modal fusion. The proposal-free method directly predicts video moments that match query sentences. Specifically, the regression-based method ~\cite{yuan2019find,chen2020rethinking,lu2019debug,DRNzeng2020dense} calculates the error of time pair with ground truth for model optimization. Span-based method ~\cite{zhao2021cascaded,VSLNet_Lzhang2021natural} predicts the probabilities of each video frame being the starting, ending and content location of the target moment. 
Existing methods ignore the annotation sparsity in video grounding, DRN~\cite{DRNzeng2020dense} is the pioneer to notice this issue which uses the distance between frames within the ground truth and the starting (ending) frame as dense supervision signals. However, DRN does not exploit moment-sentence pairs of unannotated video frames. In this paper, we leverage potential information in them to significantly improve the grounding performance.

\noindent \textbf{Dense Video Captioning.} Dense video captioning~\cite{krishna2017dense,li2018jointly,suin2020efficient,yang2023concept,mao2023improving} techniques typically consist of event detection and caption generation. Most approaches enrich event representations through contextual modeling~\cite{wang2018bidirectional}, event-level relationships~\cite{wang2020event}, or multimodal fusion ~\cite{iashin2020multi,iashin2020better}. \cite{PDVCwang2021end} proposed a simple yet effective framework for end-to-end dense video captioning with parallel decoding (PDVC). In practice, by stacking a newly proposed event counter on the top of a transformer decoder, the ~\cite{PDVCwang2021end} precisely segments the video into several event pieces under the holistic understanding of the video content. In this work, we introduce PDVC~\cite{PDVCwang2021end} to generate dense video captions.

\section{Method}
\subsection{Problem Formulation}
Given an untrimmed video and a query sentence, we aim to retrieve a video moment that best matches the query sentence, \ie, the start time $t^s$ and end time $t^e$. We denote the video as $V=\{x_i\}^{T}_{i=1}$ frame by frame, where $x_i$ is the $i$-th frame in the video and $T$ is the total number of frames. We also represent the given sentence query as $Q=\{w_i\}^{N_q}_{i=1}$ word-by-word, where $w_i$ is the $i$-th word and $N_q$ is the total number of words.
\subsection{Feature Encoder}

\noindent \textbf{Video encoder.} We extract visual representations from the given video and encode them into a 2D temporal moment feature map following~\cite{2D-TANzhang2020learning,MMNwang2022negative,cao2022deep}. We first segment the input video into small video clips and then perform fixed interval sampling to obtain $N_v$ video clips $V=\{v_i\}^{N_v}_{i=1}$. We extract visual features using a pre-trained CNN model (\eg, C3D) and fed them into the convolutional neural network and average pooling to reduce their dimensions. Then, We construct 2D  visual feature maps $F_v\in\mathbb{R}^{N_v\times{N_v}\times{d_v}} $ referring to previous works~\cite{2D-TANzhang2020learning,MMNwang2022negative} based on visual features by max pooling and $L$ layer convolution with kernel size $K$, where $N_v$ and $d_v$ are the numbers of sampled clips and feature dimension, respectively.

\noindent \textbf{Language encoder.} Most of the existing works employ glove embedding with manually designed LSTM as the language encoder~\cite{CTRLgao2017tall,2D-TANzhang2020learning}, instead of uniformly employing pre-trained models for encoding as in the case of video processing. For a given query sentence, we generate tokens for the words $Q$ by the tokenizer and then feed them into pre-trained BERT~\cite{kenton2019bert} with text aggregation to get sentence embedding $F_q\in\mathbb{R}^{1\times d_s}$, where $d_s$ is the feature dimension.

\noindent \textbf{Unified visual-language feature embedding.} We apply two parallel convolutional or linear layers after the encoders to project $F_v$ and $F_q$ to the same feature dimension $d_n$ and employ them for regression $(V_{r}, Q_{r})$ and cross-modal contrastive learning $(V_{c}, Q_{c})$, respectively.

\begin{algorithm}[t]  
	\caption{Non-Auxiliary Caption Suppression (NACS)}
	\KwIn{$\mathcal{E}$ = [$e_1$,..., $e_M$], $e_i$ = $(s_i,t_i)$,\\ 
	$\mathcal{C}$ = [$c_1$,..., $c_M$],\\ $l_c$, $\theta$, $\mathcal{F}$}
	$\mathcal{E}$ is the set of generated moment-sentence pairs\\
    $\mathcal{C}$ contains the corresponding confidence scores\\
    $l_c$ and $\theta$ are predefined values\\
    $\mathcal{F}$ records the annotated video intervals\\
    
	\KwOut{$\mathcal{U} \leftarrow \{\}$}
	\Begin{
	    \While{$\mathcal{E} \neq$ empty {\rm and} $\mathcal{U}$.length $< l_c$}{
	        $m  \leftarrow $ argmax\:$\mathcal{C}$ \;
	        $\mathcal{U} \leftarrow \mathcal{U} \cup {e_m}$;\:$\mathcal{E}\leftarrow \mathcal{E} - e_m$\;
	        $\mathcal{C}\leftarrow \mathcal{C} - c_m$;\:$\mathcal{F}\leftarrow \mathcal{F} \cup t_m$\;

	        \For{$e_i$ in $\mathcal{E}$}{
	            $c_i \leftarrow exp(- \frac{IoU(\mathcal{F},t_i)^2}{\theta})c_i, \forall t_i\notin \mathcal{U}$\;

	        }
	    }
	    \Return $\mathcal{U}$
	}
	\label{alg1}
\end{algorithm}

\subsection{Auxiliary Caption Generation} 
In general, queries in video grounding should be visually based on the temporal region, but the boundaries of the generated dense captions are rough. Moreover, due to the complexity of the video content, there are overlaps between the dense caption intervals and the ground truth intervals. The incorrect descriptions of ground truth are detrimental to the learning of the model. To solve the above issues, we propose to exploit a reliable moment-sentence pair from the generated dense caption, \ie auxiliary caption.

Specifically, we first feed the input video into an off-the-shelf dense video captioning model (\ie PDVC~\cite{PDVCwang2021end}) to generate the dense caption set $E= \{s_i, t_i,c^s_i,c^p_i\}^M_{i=1}$, where $s_i$ and $t_i$ are the generated descriptions and corresponding timestamps, respectively; $c^s_i$ and $c^p_i$ are the confidence scores of sentences and proposals, respectively; $M$ is the pre-defined number of dense captions per video. Then, we propose Non-Auxiliary Caption Suppression (NACS) inspired by~\cite{soft-NMSbodla2017soft} for set $E$. The computation process is shown in Algorithm~\ref{alg1}. To minimize the interval overlap between auxiliary captions and between auxiliary captions and ground truth, we define $\mathcal{F}$ to record the intervals that the video is currently annotated with, which is initialized to all ground truth intervals. We calculate the confidence scores $\mathcal{C}$ and sort $\mathcal{E}$ in descending order accordingly by $\mathcal{C}$. For each $e_i$, its confidence score $c_i$ is defined as follows:

\begin{equation}
    c_i=(c^s_i + c^p_i)\frac{t^e_i - t^s_i}{d_i}
\end{equation}
where $t^s_i$ and $t^e_i$ are the start and end timestamps, respectively, and $d_i$ is the duration of the whole video. The action described by $e_i$ is considered a general action if it has a long duration, and is given a higher score. Then, the $e_i$ with the highest $c_i$ is selected and the annotated video interval $\mathcal{F}$ is updated. Finally, the confidence scores $c_i$ of other $e_i$ are decayed with a Gaussian penalty function~\cite{soft-NMSbodla2017soft} according to the degree of overlap with $\mathcal{F}$. The above operations are repeated until $\mathcal{E}$ is empty or the number of elements in $\mathcal{U}$ is equal to $l_c$.
Finally, as with the query sentence, sentences of auxiliary captions are encoded as $Q_c^t$ and $Q_r^t$ for two branches, respectively. We refer to 2D-TAN~\cite{2D-TANzhang2020learning} to encode timestamps of auxiliary captions as two-dimensional temporal maps $F_t\in\mathbb{R}^{l_c\times{N_v}\times{N_v}}$, where $l_c$ is the number of auxiliary captions. We provide details of the 2D temporal map in the supplementary material.

\subsection{Caption Guided Attention (CGA)}
\begin{figure}[t]
  \centering
   \includegraphics[width=0.9\linewidth]{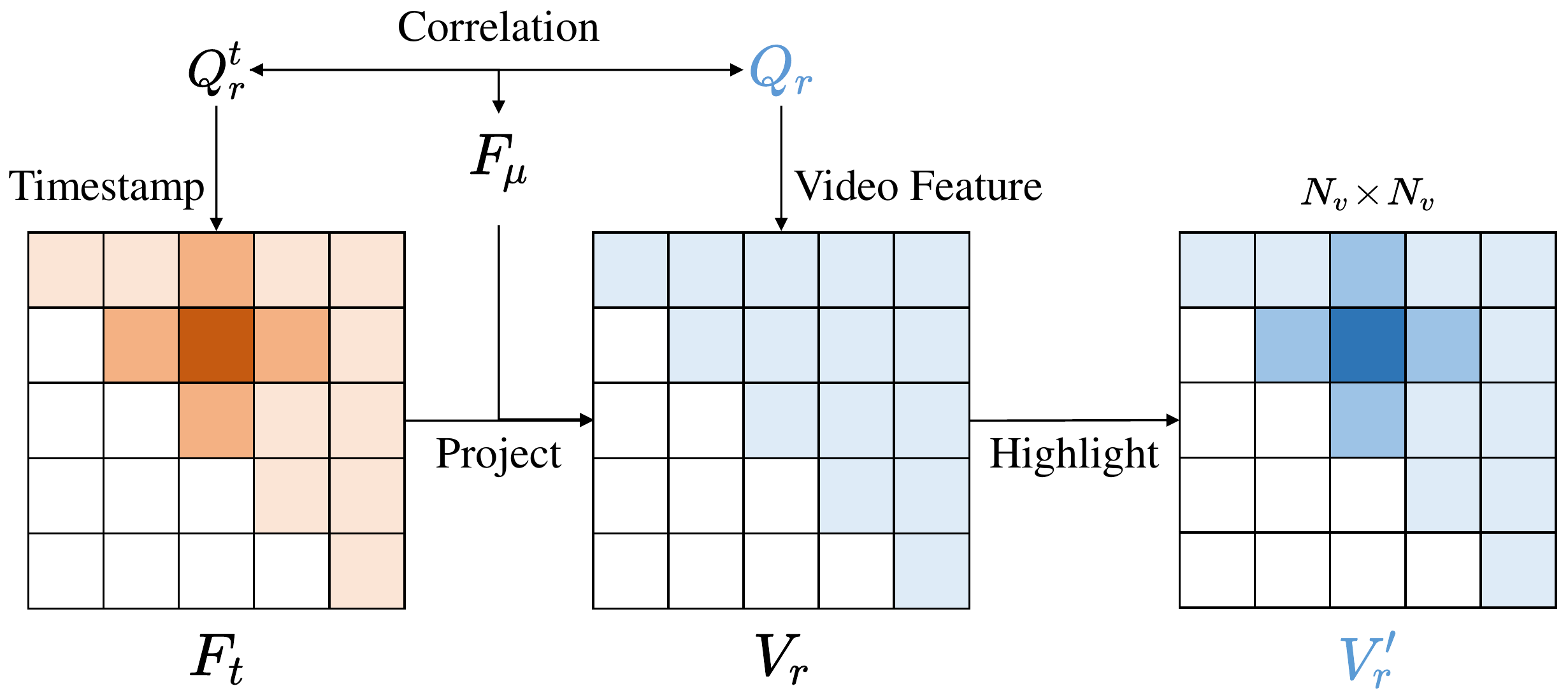}
   \caption{Illustration of our Caption Guided Attention (CGA).}
   \label{fig:project}
\end{figure}

The CGA is responsible for extracting the prior knowledge and coarse-grained estimation about the target moment from the auxiliary caption as shown in Figure~\ref{fig:project}. We first employ the co-attention mechanism to obtain the semantic relations $F_\mu$ between the sentence features $Q_r^t$ of auxiliary caption and the query sentence features $Q_{r}$:
\begin{equation}
    F_{\mu}=\operatorname{MHA}(Q_r^t, Q_r, Q_r)
\end{equation}
where MHA stands for standard multi-head attention~\cite{vaswani2017attention} which consists of $m$ parallel heads and each head is defined as scaled dot-product attention:
\begin{align}
\!\operatorname{Att}_{i}(X, Y)=\operatorname{softmax}\left(\frac{X \mathrm{W}_{i}^{\mathrm{Q}}\left(Y \mathrm{W}_{i}^{\mathrm{K}}\right)^{T}}{\sqrt{d_{m}}}\right) \!Y\!\mathrm{W}_{i}^{\mathrm{V}}\! \\
\!\operatorname{MHA}(X, Y)=\left[\operatorname{Att}_{1}(X, Y) ; \ldots ; \operatorname{Att}_{n}(X, Y)\right] \!\mathrm{W}^{\mathrm{O}}\!
\end{align}
where  $X \in \mathbb{R}^{l_{x} \times d}$  and  $Y \in \mathbb{R}^{l_{y} \times d} $ denote the Query matrix and the Key/Value matrix, respectively;  $\mathrm{W}_{i}^{\mathrm{Q}}$, $\mathrm{W}_{i}^{\mathrm{K}}$, $\mathrm{W}_{i}^{\mathrm{V}} \in   \mathbb{R}^{d \times d_{n}}$  and  $\mathrm{W}^{\mathrm{O}} \in \mathbb{R}^{d \times d}$  are learnable parameters, where  $d_{m}=d / m$. $[\cdot ; \cdot]$  stands for concatenation operation.

Then, we linearly project the semantic relation feature $F_\mu$ onto the two-dimensional temporal map $F_t$ to obtain prior knowledge $V_\mu$:
\begin{equation}
    V_\mu=\mathrm{MLP}(F_\mu \otimes F_t)
\end{equation}
where $\otimes$ represents the matrix multiplication. Note that the value in the temporal map $F_t$ represents the intersection over union (IoU), \ie temporal correlation, between the current clip and the corresponding clip of $Q_r$. Finally, we obtain $V^{\prime}_r$ used to predict the target moment by integrating prior knowledge $V_\mu$ to visual features $V_r$ by a fully connected feed-forward network:
\begin{equation}
    V^{\prime}_r=\max\left(0, (V_r+V_\mu)\mathrm{W}_{\mathrm{f}}+\mathrm{b}_{\mathrm{f}}\right)\mathrm{W}_{\mathrm{ff}}+\mathrm{b}_{\mathrm{ff}}
\end{equation}
where $\max(0,*)$ represents the ReLU activation function; $\mathrm{W}_{\mathrm{f}}$ and $\mathrm{W}_{\mathrm{ff}}$ denote learnable matrices for linear transformation; $\mathrm{b}_{\mathrm{f}}$ and $\mathrm{b}_{\mathrm{ff}}$ represent the bias terms. In this way, we assign different weights to the video features according to the semantic and temporal position of the auxiliary caption,

\subsection{Asymmetric Cross-modal Contrastive Learning (ACCL)}

\begin{figure}[t]
  \centering
   \includegraphics[width=0.9\linewidth]{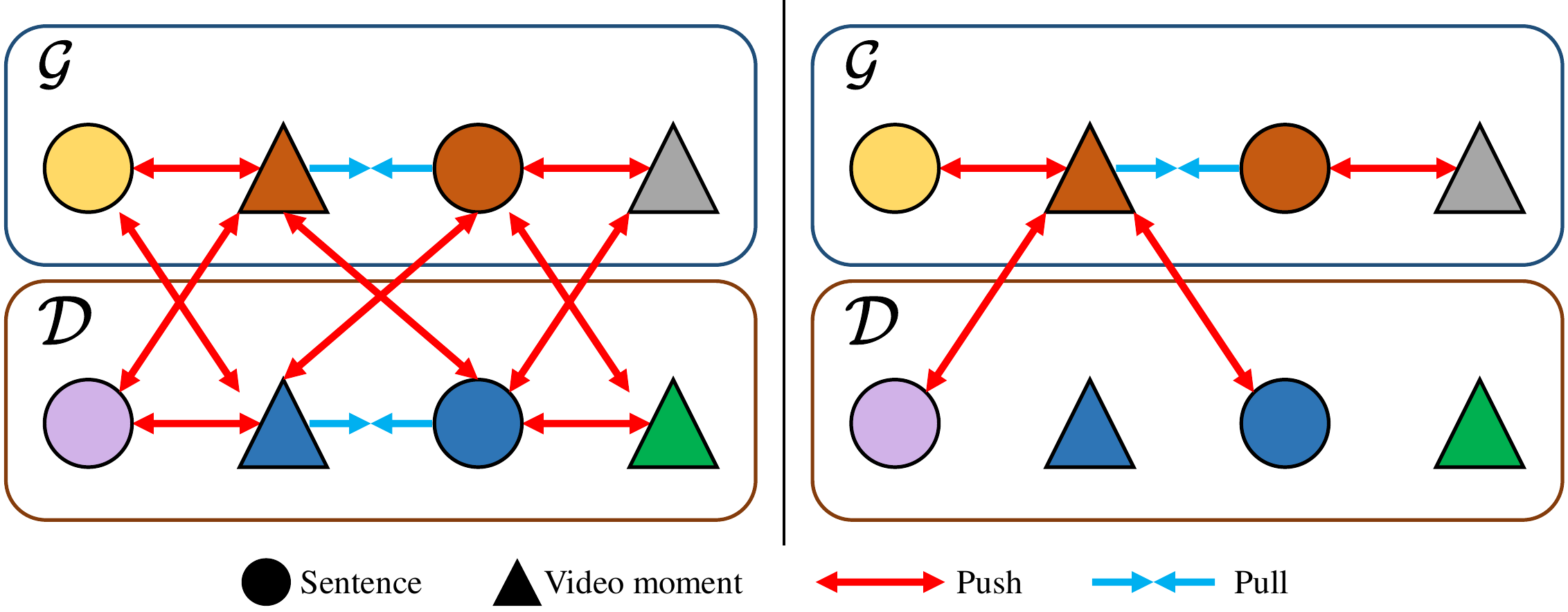}
   \caption{Illustration of our asymmetric push-and-pull strategy, in contrast to those in the original supervised contrastive learning, where elements with the same color mean they come from the same moment-sentence pair. $\mathcal{G}$ and $\mathcal{D}$ are the sets of moment-sentence pairs of ground truth and auxiliary caption, respectively.}
   \label{fig:intro_cl}
\end{figure}

For traditional cross-modal contrastive learning, matched pairs are considered as positive pairs and mismatched pairs are considered as negative pairs. However, the temporal boundaries of auxiliary caption may be coarse and should not be pulled close to the corresponding sentences for the localization task. In addition, the intervals of auxiliary caption may overlap with ground truth so as to disagree on the same moment. Therefore, we propose asymmetric cross-modal contrastive learning (ACCL). We consider video grounding as a dual matching task, \ie moment to text and text to moment. Figure~\ref{fig:intro_cl} illustrates the core idea of ACCL: ACCL applies pulling and pushing in ground truth pairs, and applies pushing between ground truth moments and prompt sentences. We adopt the noise contrastive estimation (NCE)~\cite{gutmann2010noise} to calculate our ACCL loss, which is defined as:
\begin{gather}
    \mathcal{L}_{c} = \lambda_{v}\mathcal{I}_{v\rightarrow s} + \lambda_{s}\mathcal{I}_{s\rightarrow v}\\
    \!\mathcal{I}_{v\rightarrow s}\!=\!-\frac{1}{|\mathcal{P}|} \sum_{i \in \mathcal{P}}\log\frac{\operatorname{exp}(f(v_i)\!^\top\!f(s_i)/\tau_v)} {{\sum_{j\in\mathcal{A}_{s}}\!}\operatorname{exp}(f(v_i)\!^\top\!f(s_j)/\tau _v)}\!
    \label{eq:cl vs}\\
    \!\mathcal{I}_{s\rightarrow v}\!=\!-\frac{1}{|\mathcal{P}|} \sum_{i \in \mathcal{P}} \log\frac{\operatorname{exp}(f(s_i)\!^\top\!f(v_i)/\tau_s)}{{\sum_{j\in\mathcal{A}_{v}}\!}\operatorname{exp}(f(s_i)\!^\top\!f(v_j)/\tau _s)}\!
    \label{eq:cl sv}
\end{gather} 
where $i$ and $j$ are indexes of video moment $v$ or sentence $s$ from $V_c$, $Q_c$ and $Q_c^t$; $\lambda_{v}$ and $\lambda_{s}$ are hyperparameters to balance the contribution of contrastive loss for each direction; $\tau_v$ and $\tau_s$ are temperatures. At first glance, $\mathcal{I}_{v\rightarrow s}$ and $\mathcal{I}_{s\rightarrow v}$ \textit{seem} identical to the vanilla cross-modal contrastive learning loss. However, the \textit{key difference} lies in the definitions of $\mathcal{P}$, $\mathcal{A}_s$ and $\mathcal{A}_v$, as we detail below.

\textbf{Asymmetry of Positive pairs and Negative pairs (APN).} We do not pull moments of the auxiliary caption and sentences together, \ie, $\mathcal{P}=\mathcal{G}$. This design is motivated by the fact that we cannot guarantee the accuracy of the auxiliary caption. The boundaries of the auxiliary caption moments are rough, while video grounding is an exact and frame-level matching task. If we construct them as positive pairs, which will hinder cross-modal learning for exact matching.

\textbf{Asymmetry of Negative pairs in Dual Matching (ANDM).} Moment-sentence pairs of auxiliary caption are only contained in $\mathcal{A}_{s}$ and not in $\mathcal{A}_{v}$, \ie, $\mathcal{A}_{s}=\mathcal{G}_s \cup \mathcal{D}_s$, $\mathcal{A}_{v}=\mathcal{G}_v$. We only push target moments away from auxiliary caption sentences, but do not push query sentences away from auxiliary caption moments. Since auxiliary caption moments and target moments are independent of each other and they may refer to the same video moments, \ie, it is possible for auxiliary caption moments to match with query sentences. On the other hand, the auxiliary caption sentences provide more descriptions of the video content, and we treat the moment-sentence pairs they form with the ground truth moments as hard negative pairs to enhance joint representation learning.

\subsection{Training and Inference}
\textbf{Training.} In the regression branch, we employ cross-entropy loss to optimize the model: 
\begin{equation}
    \mathcal{L}_{r} = \frac{1}{C} \sum_{i=1}^{C} y_{i} \log p_{i}+\left(1-y_{i}\right) \log \left(1-p_{i}\right)
    \label{eq:bce}
\end{equation}
where $p_i$ is the prediction score of a moment and $C$ is the total number of valid candidates. Our contrastive loss relies on the binary supervision signal to learn cross-modal alignment and the regression loss counts on the IoU supervision signal for moment ranking. Finally, we employ these two complementary losses for training.
The overall training loss $\mathcal{L}$ of our model is
\begin{equation}
    \mathcal{L} = \lambda_c\mathcal{L}_{c} + \lambda_r\mathcal{L}_{r}
\end{equation}
where $\lambda_c$ and $\lambda_b$ are hyperparameters to balance the contribution of each loss.

\textbf{Inference.} During inference, we calculate the cosine similarity of the video moments and queries as the prediction scores
\begin{equation}
    \mathcal{S}_r = \sigma(f(Q_r)f(V^{\prime}_r)^\top),\; \mathcal{S}_c = f(Q_c)f(V_c)^\top
\end{equation}
where $\sigma$ is the sigmoid function. 

Due to the difference in the value region of $\mathcal{S}_{r}$ and $\mathcal{S}_{c}$ (especially the negative regions), we fuse them after scaling to obtain the final prediction scores $\mathcal{S}$.
\begin{equation}
    \mathcal{S} = \mathcal{S}_{r}\odot(\frac{\mathcal{S}_{c}+1}{2})^{\gamma}
\end{equation}
where $\odot$ denote the element-wise multiplication and $\gamma$ is the hyperparameter. 
Finally, We rank all the moment proposals according to $\mathcal{S}$ followed by a non-maximum suppression (NMS) function.
\begin{table*}[t]
\centering
\footnotesize
\centering
\begin{tabular}{l|cccccc|cccccc}
\toprule
\multirow{3}{*}{Method} & \multicolumn{6}{c|}{ActivityNet Captions}            & \multicolumn{6}{c}{TACoS}                           \\ \cmidrule{2-13} 
                        & R@1    & R@1    & R@1    & R@5    & R@5    & R@5    & R@1    & R@1    & R@1    & R@5    & R@5    & R@5    \\
                        & IoU0.3 & IoU0.5 & IoU0.7 & IoU0.3 & IoU0.5 & IoU0.7 & IoU0.1 & IoU0.3 & IoU0.5 & IoU0.1 & IoU0.3 & IoU0.5 \\ \cmidrule{1-13}
CTRL              & 47.43  & 29.01  & 10.34  & 75.32  & 59.17  & 37.54  & 24.32  & 18.32  & 13.30  & 48.73  & 36.69  & 25.42  \\
CBP        & 54.30  & 35.76  & 17.80  & 77.63  & 65.89  & 46.20  & --     & 27.31  & 24.79  & --     & 43.64  & 37.40  \\
SCDM         & 54.80  & 36.75  & 19.86  & 77.29  & 64.99  & 41.53  & --     & 26.11  & 21.17  & --     & 40.16  & 32.18  \\
2D-TAN    & 59.45  & 44.51  & 26.54  & 85.53  & 77.13  & 61.96  & 47.59  & 37.29  & 25.32  & 70.31  & 57.81  & 45.04  \\
DRN              & --     & 45.45  & 24.39  & --     & 77.97  & 50.30  & --     & --     & 23.17  & --     & --     & 33.36  \\
FVMR              & 60.63  & 45.00  & 26.85  & 86.11  & 77.42  & 61.04  & {53.12}  & 41.48  & 29.12  & {78.12}  & 64.53  & 50.00  \\
RaNet        & --     & 45.59  & 28.67  & --     & 75.93  & 62.97  & --     & 43.34  & 33.54  & --     & 67.33  & 55.09  \\
DPIN             & --     & 47.27  & 28.31  & --     & 77.45  & 60.03  & --     & 46.74  & 32.92  & --     & 62.16  & 50.26  \\
MATN           & --     & 48.02  & 31.78  & --     & 78.02  & 63.18  & --     & \underline{48.79}  & \underline{37.57}  & --     & {67.63}  & {57.91}  \\
CBLN           & {66.34}  & 48.12  & 27.60  & \underline{88.91}  & 79.32  & 63.41  & 49.16  & 38.98  & 27.65  & 73.12  & 59.96  & 46.24  \\
SMIN       & --     & 48.46  & 30.34  & --     & 81.16  & 62.11  & --     & 48.01  & 35.24  & --     & 65.18  & 53.36  \\
GTR             & --     & 50.57  & 29.11  & --     & 80.43  & 65.14  & --     & 40.39  & 30.22  & --     & 61.94  & 47.73  \\
MMN           & 65.05  & 48.59  & 29.26  & 87.25  & 79.50  & 64.76  & 51.39  & 39.24  & 26.17  & 78.03  & 62.03  & 47.39  \\
SPL            & --     & {52.89}  & {32.04}  & --     & \underline{82.65}  & {67.21}  & --     & 42.73  & 32.58  & --     & 64.30  & 50.17  \\
G2L & -- & {51.68} & {33.35} & -- & {81.32} & {67.60} & -- & {42.74} & {30.95} & -- & {65.83} & {49.86} \\
\midrule
ACNet  & {66.82} & {52.51} & {32.51} & {87.11} & {79.89} & {66.68} & {57.66} & {48.13} & {36.79} & {80.11} & {69.08} & {58.10}\\
ACNet$^\Diamond$ & \underline{67.07} & \underline{53.55} & \underline{34.68} & {88.21} & {80.94} & \underline{67.78} & \underline{58.76} & {48.74} & {37.14} & \underline{82.43} & \underline{71.47} & \underline{60.66}\\

ACNet$^\natural$ & \textbf{70.31} & \textbf{56.39} & \textbf{38.19} & \textbf{89.26} & \textbf{82.87} & \textbf{70.77} & \textbf{62.76} & \textbf{51.64} & \textbf{38.84} & \textbf{86.83} & \textbf{74.73} & \textbf{62.86}\\
\bottomrule
\end{tabular}
\caption{Performance comparisons on ActivityNet Captions and TACoS. $^\Diamond$ denotes using the generated auxiliary captions and $^\natural$ denotes introducing manual annotations from other moments within the video as auxiliary captions during inference.}
\label{tab:com sota}
\end{table*}

\section{Experiments}
\subsection{Datasets and Evaluation}

\textbf{ActivityNet Captions.} ActivityNet Captions~\cite{krishna2017dense} contains 20,000 untrimmed videos and 100,000 descriptions from YouTube ~\cite{caba2015activitynet}, covering a wide range of complex human behavior. The average length of the videos is 2 minutes, while video clips with annotations have much larger variations, ranging from a few seconds to over 3 minutes. Following the public split, we use 37417, 17505 and 17031 sentence-video pairs for training, validation and testing, respectively.

\textbf{TACoS.} TACoS~\cite{regneri2013grounding} contains 127 videos from the cooking scenarios, with an average of around 7 minutes. We follow the standard split ~\cite{CTRLgao2017tall}, which has 10146, 4589 and 4083 video query pairs for training, validation and testing, respectively.

\textbf{ActivityNet-CG.} ActivityNet-CG~\cite{VISAli2022compositional} aims to evaluate how well a model can generalize to query sentences that contain novel compositions or novel words. It is a new split of ActivityNet Captions, which is re-split into four sets: training, novel-composition, novel-word, and test-trivial.

\textbf{Evaluation.} Following previous work~\cite{CTRLgao2017tall,2D-TANzhang2020learning}, we adopt ``R@n, IoU=m" as the evaluation metric. It calculates the percentage of IoU greater than ``m" between at least one of the top ``n" video moments retrieved and the ground truth. 

\subsection{Implementation Details}
Following~\cite{2D-TANzhang2020learning,MMNwang2022negative}, we employed a 2D feature map to generate moment proposals. For the input video, we used exactly the same settings as in the previous work~\cite{MMNwang2022negative} for a fair comparison, including visual features (both C3D features), NMS thresholds (0.5, 0.4), number of sampled clips (64, 128), number of 2D convolution network layers (3, 4) and kernels (4, 2) for ActivityNet Captions and TACoS, respectively. For the query sentence, the pre-trained BERT~\cite{kenton2019bert} was employed for each word of the query. Specifically, the average pooling results of the last two layers are used to obtain the embedding of the whole sentence. During the training, we used AdamW~\cite{loshchilov2018decoupled} optimizer to train our model with learning rate of $8 \times 10^{-4}$. The batch size $B$ was set to 48 and 8 for ActivityNet Captions and TACoS, respectively. We employed the same settings as ActivityNet Captions on ActivityNet-CG.
\subsection{Comparison with State-of-the-art Methods}
\textbf{Benchmark.} We compare our ACNet with state-of-the-art methods in Table~\ref{tab:com sota}. ACNet achieves significant improvements compared with all other methods. Specifically, on ActivityNet Captions, our ACNet achieves performance improvements of up to 6\% compared with the cutting edge method SPL~\cite{SPLliu2022skimming}. SPL~\cite{SPLliu2022skimming} investigates the imbalance of positive and negative frames in video grounding and develops a coarse-grained and fine-grained two-step framework, but does not consider the relationship between potential actions and queries. In contrast, our method encodes the video feature under the guidance of the auxiliary caption with a stronger correlation to the query. For TACoS, our ACNet outperforms the strongest competitor MATN~\cite{MATNzhang2021multi} by up to 7 points. MATN~\cite{MATNzhang2021multi} proposes a multi-level aggregated transformer, but it can easily overfit to the point of confusing similar actions due to the neglect of the sparse annotation dilemma. Our ACNet mines more supervision signals from the unannotated moments and employs two complementary loss functions to improve the grounding quality. Notably, most methods cannot achieve the best performance on both datasets simultaneously due to the differences between the two datasets, but ACNet does, which demonstrates the superiority of our method.

\begin{table}[t]
\begin{center}
\footnotesize
\begin{tabular}{@{}lcccc@{}}
\toprule
\multirow{3}{*}{Method} & \multicolumn{2}{c}{Test-Trivial} & \multicolumn{2}{c}{Novel-Comp} 
\\ \cmidrule(l){2-5} 
 & R@1 & R@1 & R@1 & R@1 \\
 & IoU0.5 & IoU0.7 & IoU0.5 & IoU0.7 \\ \midrule
TMN & 16.82 & 7.01 & 8.74 & 4.39  \\
TSP-PRL & 34.27 & 18.80 & 14.74 & 1.43 \\
VSLNet & 39.27 & 23.12 & 20.21 & 9.18  \\
LGI & 43.56 & 23.29 & 23.21 & 9.02  \\
2D-TAN & 44.50 & 26.03 & 22.80 & 9.95  \\
VISA & {47.13} & {29.64} & {31.51} & {16.73}  \\ \midrule
ACNet & \textbf{51.81} & \textbf{33.52} & \textbf{33.30} & \textbf{17.09} \\
ACNet$^\dagger$ & 46.33 & 28.67 & 30.71 & 15.80 \\ \bottomrule
\end{tabular}
\end{center}
\caption{Performance comparison on ActivityNet-CG.  ``$\dagger$" denotes without NACS.}
\label{tab:anet-cg}
\end{table}

\begin{table}[t]
\begin{center}
\footnotesize
\begin{tabular}{cccc|ccc}
\toprule
\multirow{2}{*}{NACS} & \multirow{2}{*}{CGA} & \multirow{2}{*}{ACCL} & \multirow{2}{*}{Reg} & R@1 & R@1 & R@1 \\
 &  &  &  & \multicolumn{1}{l}{IoU0.3} & \multicolumn{1}{l}{IoU0.5} & \multicolumn{1}{l}{IoU0.7} \\ \midrule
 &  &  &  \checkmark &  62.73 & 46.74 & 27.12 \\
 &  & \checkmark &  & 64.57 & 47.28 & 28.09 \\ \midrule
  \checkmark & &  \checkmark & & 66.74 & 51.83 & 32.29 \\
 & \checkmark &  &  \checkmark & 67.70 & 52.08 & 32.02 \\ \midrule
   & &  \checkmark & \checkmark & 65.03 & 50.24 & 30.02 \\
\checkmark &  & \checkmark &  \checkmark & 68.83 & 54.85 & 36.48 \\
& \checkmark & \checkmark &  \checkmark & 68.36 & 55.27 & 36.91 \\
\midrule

\checkmark & \checkmark & \checkmark &  \checkmark & \textbf{70.31} & \textbf{56.39} & \textbf{38.19} \\
\bottomrule
\end{tabular}
\end{center}
\caption{Component ablations on ActivityNet Captions.}
\label{tab:main ablation}
\end{table}

\textbf{Compositional Generalization.} Table~\ref{tab:anet-cg} shows the result comparison between state-of-the-art methods on ActivityNet-CG. Unlike ActivityNet Captions and TACoS, ActivityNet-CG focuses on verifying the generalizability of the model on novel compositions or novel words, proposed by VISA~\cite{VISAli2022compositional}. We observe that our ACNet brings performance improvement of up to 4\% compared with VISA~\cite{VISAli2022compositional}, demonstrating the excellent compositional generalization of our model. Notably, our variant ``$\dagger$" model is weaker than VISA on all splits, indicating that auxiliary caption is crucial for generalizability.

\subsection{Ablation Study}
\textbf{Main Ablation Study.} 
In Table~\ref{tab:main ablation}, we conduct a thorough ablation study on the proposed components to verify their effectiveness. The first two rows of Table~\ref{tab:main ablation} show our single-branch base model. Based on these, we add NACS and CGA respectively. it can be noticed that the performance is improved by about 4\% and 5\% respectively, as shown in the third and fourth rows of Table~\ref{tab:main ablation}. Row 5 of Table~\ref{tab:main ablation} shows our two-branch base model, which improves ``IoU=0.5'' to 50.24. In rows 6 and 7 of Table~\ref{tab:main ablation}, we add NACS and CGA, respectively, to the two-branch model and find that the performance improves again by about 5\%. The last row of Table~\ref{tab:main ablation} shows the performance of our full model, which further improves the ``IoU=0.5'' to 56.39\% and achieves the best performance among ablation variants.

\begin{table}[t]
\footnotesize
\centering
\begin{tabular}{lcc}
\toprule
Model & Training & Inference \\
\midrule
2D-TAN~\cite{2D-TANzhang2020learning} & 0.13s & 32s \\
MMN~\cite{MMNwang2022negative} & 0.32s &  37s \\
\midrule
Base Model & 0.39s & 40s \\
\textbf{ACNet} & 0.94s & 53s \\
\bottomrule
\end{tabular}
\captionof{table}{Time consumption on ActivityNet-Captions.}
\label{tab: abl cost}
\end{table}
\hfill
\begin{table}[t]
\footnotesize
\centering
\begin{tabular}{lccc}
\toprule
Model  & R@1 0.3 & R@1 0.5 & R@1 0.7 \\ \midrule
CL & 66.25 & 48.59 & 30.34 \\
w/o APN  & 67.43& 53.79 & 37.68 \\
w/o ANDM  & 67.54 & 53.58 & 37.70 \\ \midrule
Full ACCL & 70.31 & 56.39 & 38.19 \\
\bottomrule
\end{tabular}
\captionof{table}{ Ablation studies of ACCL on ActivityNet Captions.}
\label{tab: abl accl}
\end{table}

\textbf{Comparisons of Time Consumption.} In Table~\ref{tab: abl cost}, we compute the average training time per iteration and total inference time. Our method requires more computational costs but these are worth compared to the significant performance improvements.

\textbf{Effect of Asymmetric Components.} To evaluate the detailed components in ACCL more deeply, we conduct an ablation study of APN and ANDM on ActivityNet Captions in Table~\ref{tab: abl accl}. We observe that removing any of the components brings significant performance degradation, indicating that this asymmetric design is capable of mining more hard negative samples from the auxiliary caption and thus improving the representation learning.


\section{Conclusion}
In this paper, we propose an Auxiliary Caption Network (ACNet) for video grounding. Firstly, we propose Non-Auxiliary Caption Suppression (NACS) to obtain auxiliary captions from dense captions. Then, we design a simple but effective Caption Guided Attention (CGA) to extract prior knowledge from the auxiliary captions and approximately locate the target moment. Moreover, we propose Asymmetric Cross-modal Contrastive Learning  (ACCL) to fully mine negative pairs and construct extra supervision signals from unannotated video clips. Extensive experiments have demonstrated that ACNet can achieve excellent performance and superior generalizability on public datasets.


\section{Acknowledgments}
This paper was partially supported by NSFC (No: 62176008) and Shenzhen Science \& Technology Research Program (No: GXWD20201231165807007-20200814115301001).

\bibliography{aaai24}

\end{document}